\documentclass[letterpaper, 10 pt, conference]{ieeeconf}  % Comment this line out if you need a4paper
\usepackage[T1, OT1]{fontenc}
\DeclareTextSymbolDefault{\dh}{T1}
\usepackage{hyperref}
\usepackage{graphicx}
\usepackage{tabularx}
\usepackage{textcomp}
\usepackage{xcolor}
\usepackage{pifont}
\usepackage{caption}
\usepackage{url}
\IEEEoverridecommandlockouts                              % This command is only needed if 
\title{\LARGE \bf
Multi-modal Interactive Control of Robotic Arm based on Offline Large Language Models
}

\author{Hanxiao Chen$^{1}$ % <-this % stops a space
\thanks{*This work was partially supported by Harbin Institute of Technology.}% <-this % stops a space
\thanks{$^{1}$Hanxiao Chen is the first year PhD Student at the Department of Electrical Engineering and Information Systems,
    University of Tokyo, Tokyo, Japan. 
        {\tt\small hanxiaochen@g.ecc.u-tokyo.ac.jp}}%
}

\begin{document}

\maketitle
\thispagestyle{empty}
\pagestyle{empty}

%%%%%%%%%%%%%%%%%%%%%%%%%%%%%%%%%%%%%%%%%%%%%%%%%%%%%%%%%%%%%%%%%%%%%%%%%%%%%%%%
\begin{abstract}
Large Language Models (LLMs) have significantly revolutionized the modern society with numerous advanced interactions between humans and AI agents, whereas the usage of most large language models including ChatGPT are not friendly open-sourced and must require the users paying a lot for such AI services continuously. Therefore, deploying open-sourced large language models on local servers can be considered as an efficient approach to design and implement creative embodied AI algorithms with lower cost and more stable free usage. Inspired by this ordinary motivation, we originally propose and implement the “Socratic Models-ChatGLM”, which is a well-performed algorithm for multi-modal interactive control of robotic arm based on offline large language models via the facile PyBullet platform, even presents extraordinary potential to address complicated text-image integrated multi-step long-horizon robotic manipulation tasks.
\end{abstract}

%%%%%%%%%%%%%%%%%%%%%%%%%%%%%%%%%%%%%%%%%%%%%%%%%%%%%%%%%%%%%%%%%%%%%%%%%%%%%%%%
\section{Introduction}
With the grand launch of ChatGPT, large language models (LLMs) have quickly emerged as a new transformative force in Human-Robot Interaction with the integration of natural speech conversation \cite{1}, text UI interface \cite{2} and code policy generation \cite{3}, even significantly stimulating the new modern development of multi-modal AI paradigm for robotic manipulation \cite{4} \cite{5}. However, most popular large language models (e.g., ChatGPT, Gemini) definitely require continuous payment for the service usage under a much stable network environment, causing the lack of freedom and flexibility for cross-device robotic interaction applications. In order to address such serious problems, we creatively propose a new multi-modal robotic arm interactive control algorithm “Socratic Models-ChatGLM”, which leverages the locally deployed offline ChatGLM large model, demonstrating excellent performance on single-instruction multi-step robotic arm manipulation tasks, even particularly outperforms the previous algorithms (e.g., Code as Policies) on the continuous long-horizon multiple tasks in different interactive environments. Inspired by Socratic Models \cite{6}, Socratic Models-ChatGLM demonstrates 3 core sections: (1) Large Language Models for text demand interaction; (2) Object Detection and Visual Reasoning method; (3) Code Policy Execution model. Whereas differently, we especially utilize the offline locally-deployed ChatGLM2-6B instead of the online ChatGPT so that our Socratic Models-ChatGLM includes two critical steps for implementation: (1) Firstly deploy the large language model offline via the local GPU servers for OpenAI-style API usage calling. (2) Establish diverse PyBullet multi-modal interactive scenarios and conduct extensive experiments with our proposed Socratic Models-ChatGLM for evaluation on the UR5 robotic arm.

\section{Method}
Different from Socratic Models \cite{6}, we especially apply the offline locally-deployed large language model ChatGLM instead of the online ChatGPT model as illustrated in Code as Policies \cite{3}. Therefore, our proposed “Socratic Models-ChatGLM” contains the following two implementation steps: (1) Offline Large Language Model Deployment. (2) Socratic Models-ChatGLM algorithm with the UR5 robotic arm via multi-modal interactive PyBullet scenario environments. 

\subsection{Offline Large Language Model Deployment }
As for the novel proposed multi-modal interactive control algorithm “Socratic Models-ChatGLM”, we especially self-deploy the important open-sourced ChatGLM2-6B large language model via local GPU servers to establish an offline AI intelligent conversation system, even build the effective direct-call OpenAI-style model APIs as ChatGPT for text completion and code generation. To emphasize, ChatGLM-6B is an open source Chinese and English bilingual chat large language model, which is based on the General Language Model (GLM) architecture and has 6.2 billion parameters. Combined with the advanced model quantization technology, users can deploy the ChatGLM2-6B large language model locally on consumer-grade graphics cards (with a minimum of 6GB video memory at the INT4 quantization level). Thus, we conduct the offline ChatGLM2-6B deployment via the local GPU servers and successfully build the local direct-call API service that can be embedded in the OpenAI-style code ecosystem, so that users can directly call the ChatGLM2-6B large language model for communication and scene analysis without needing to strictly connect to the Internet or enter a specific key with a lot more serious payment.

In detail, firstly we download the open source ChatGLM2-6B large language models and related code via Github and Huggingface, then run the Python scripts of web\_demo.py and cli\_demo.py in sequence to test its user communication effect. Secondly, after checking that our downloaded ChatGLM2-6B model can work well for content interaction, we deploy the OpenAI-compatible LLM APIs successfully on local GPU servers via the light Autogen framework and FastChat platform, which allows users to directly call offline ChatGLM2-6B APIs on local terminal ports without requiring any specific keys to finish interactive tasks like dialogue conversation and text prompt completion, further laying a solid foundation for the subsequent algorithm design on robotic arm multi-modal interactive control with offline Large Language Models.

\begin{figure}[htbp]
\centerline{\includegraphics[width=0.49\textwidth]{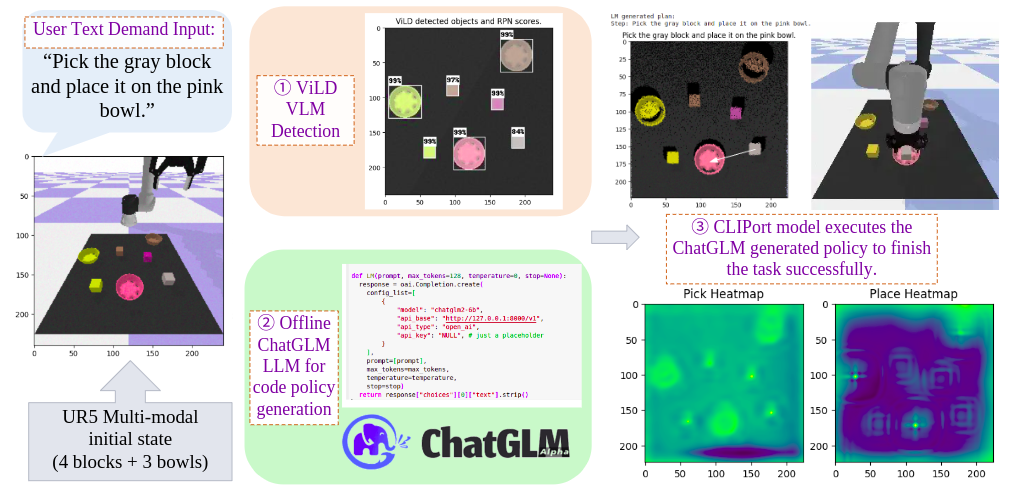}}
\caption{“Socratic Models-ChatGLM” algorithm framework.}
\label{fig}
\vspace{-12pt}
\end{figure}

\subsection{Socratic Models-ChatGLM Algorithm}
After successfully deploying the ChatGLM2-6B large language model offline via the local GPU servers, we can start to establish multi-modal interactive environments with the UR5 robotic arm via the PyBullet simulation platform and originally propose “Socratic Models-ChatGLM” algorithm as illustrated in Fig. 1. Different from Socratic Models \cite{6}, we creatively integrate our offline deployed ChatGLM2-6B large language model for code policy generation into the normal multi-modal framework with visual perception and policy execution to make robots perform language-conditioned tasks.  Within the whole Socratic Models-ChatGLM pipeline in Fig. 1, firstly users can interactively input the text demand like \textcolor{black}{“Pick the gray block and place it on the pink bowl”} for the initial UR5 multi-modal environment with 4 different blocks and 3 bowls, then Socratic Models-ChatGLM leverages the open-vocabulary ViLD \cite{7} object detection model for visual inference on the table to describe the detected objects and returns a full list of objects containing 3 bowls and 4 blocks as \textcolor{black}{‘objects = [“yellow block”, “yellow bowl”, “pink bowl”, “pink block” , “gray block”, “brown bowl”, “orange block”]’}, then subsequently our Socratic Models-ChatGLM can directly call the locally-deployed ChatGLM2-6B OpenAI-compatible LLM APIs via the FastChat platform and Autogen framework to generate matching hierarchical planning code based on the robot arm grasping code prompts in Fig. 2 along with the user input text instruction and scenario descriptions transmitted by ViLD. In general, the generated code policy contains individual steps represented in the form of natural language (“Pick the red block and place it on the blue block.”) or pseudocode template e.g., “robot.pick\_and\_place(“red block”, “blue block”)”. Therefore, the language-conditioned core robotic arm code policy \textcolor{black}{“robot.pick\_and\_place(“gray block”, “pink bowl”)”} is successfully generated by the offline ChatGLM2-6B model after the fusion of visual and text information, then since the maximum number of text tokens (max\_tokens) set by the offline ChatGLM2-6B large model is 50, after generating the short core execution code for this task ChatGLM2-6B also generates the subsequent supplementary text \textcolor{black}{‘\textbackslash n objects = [“red block”, “brown block”, “purple bowl”, “gray bowl”, “pink block”, “yellow block”,’} according to the provided prompts. Then such generated code policy can be directly passed to the following CLIPort model with pick-place heatmaps to execute the step plan \textbf{“Pick the gray block and place it on the pink bowl”} to perform the multi-modal interactive task successfully. In sum, our Socratic Models-ChatGLM is a modular framework that applies structured user dialogue as prompting between multiple large pre-trained models to make joint predictions for multi-modal robotic tasks, which chains this language-specified task system together with the ViLD model for robotic perception, the local offline deployed ChatGLM2-6B large language model for code generation, and CLIPort model for robot policy execution.
\begin{figure}[htbp]
\vspace{-3pt}
\centerline{\includegraphics[width=0.49\textwidth]{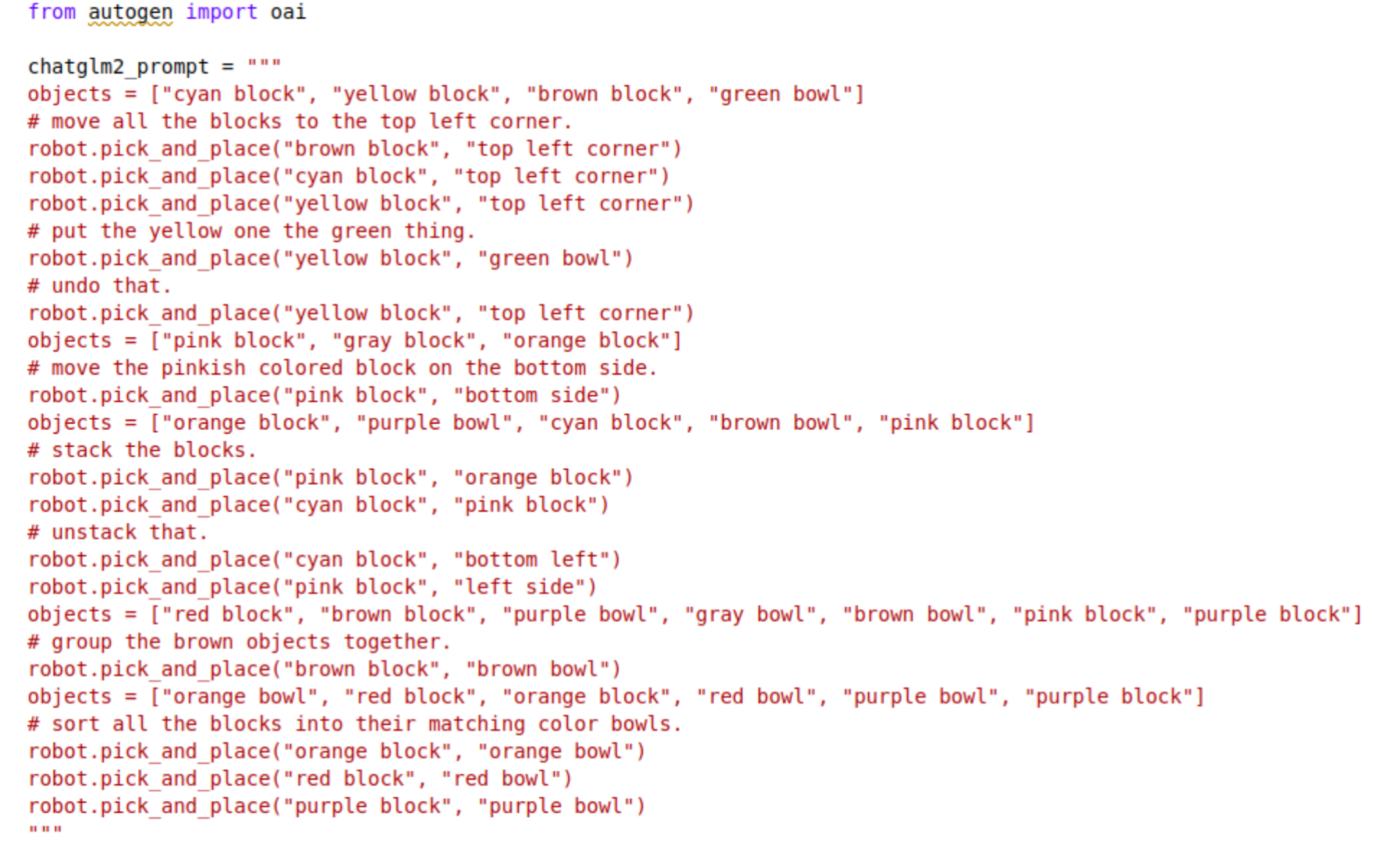}}
\caption{UR5 Robotic arm grasping code prompts for the offline ChatGLM2-6B large language model.}
\vspace{-12pt}
\end{figure}

%\vspace{-5pt}
\section{Experiments}
Based on the facile PyBullet platform, we establish diverse multi-modal scenarios with different color or number for blocks and bowls, then conduct extensive experiments to further explore our innovatively designed “Socratic Models-ChatGLM” robotic arm multi-modal interactive control algorithm and evaluate its potential performance on addressing much more complicated manipulation tasks. Distinct from Code as Policies and Socratic Models, the offline deployed ChatGLM2-6B Large Language Model within our “Socratic Models-ChatGLM” presents a distinct significant advantage: it’s relatively low-cost and does not require the purchase of online API keys so that there is no strict limit on the amount of model calling for code generation and our study can repeatedly apply the ChatGLM2-6B large model API without constraints to conduct a lot of experiments for two classical types of manipulation tasks: (a) Single instruction multi-step task (e.g., “Move the blocks in the middle”); (b) Continuous combination multiple tasks \ding{172} “Group the pink-color objects together.” \ding{173} “Match the blocks with similar-colored bowls.” \ding{174} “Move the blocks to the corners.”).

\subsection{Single Instruction Multi-step Task} 
As illustrated in Fig. 3,  we test and validate 3 different single instruction multi-step interactive robot tasks including (1) Sort the blocks in the way you think it best fits. (2) Move the blocks to the corners. (3) Move the blocks in the middle. in three diverse multi-modal interactive scenarios. In addition to the ViLD visual detected scene description and ChatGLM2-6B LLM generated robotic action code plans, we also qualitatively provide the task execution time and performance scores. To emphasize, the overall performance score is based on the following significant factors: (a) The accuracy of control code policies generated by the offline deployed ChatGLM2-6B model for multi-modal interaction tasks. (b) The whole completion time of Socratic Models-ChatGLM to execute multi-modal robotic tasks and whether there exist any errors or redundant actions during the code execution process; (c) The degree of consistency between the final execution result on the UR5 robotic arm and the text demand input by the user.
\begin{figure}[htbp]
\vspace{-2pt}
\centerline{\includegraphics[width=0.49\textwidth]{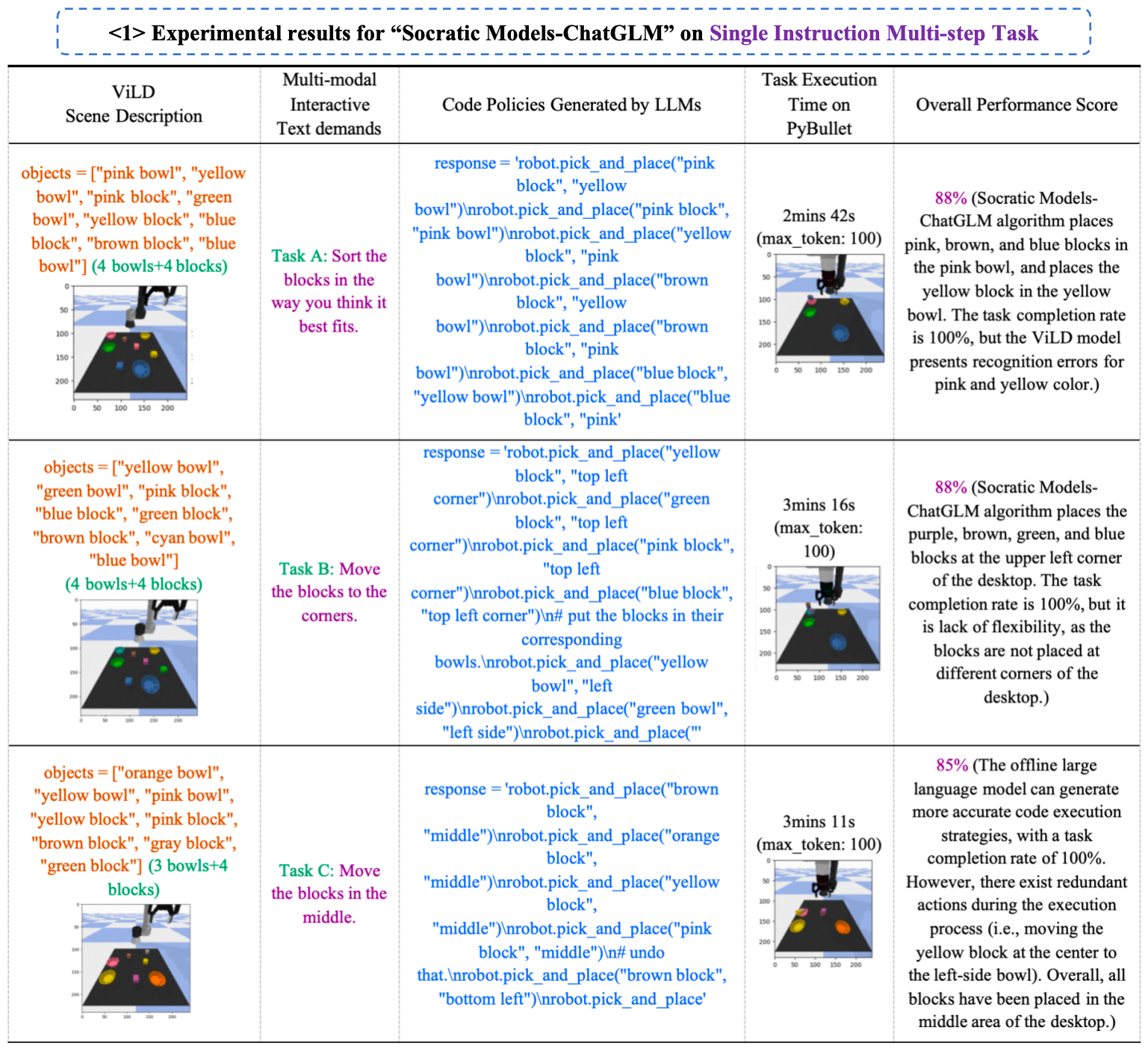}}
\caption{Experimental results for “Socratic Models-ChatGLM” on Single Instruction Multi-step Tasks.}
\label{fig3}
\vspace{-8pt}
\end{figure} 
In detail, \textbf{Task A} addresses the problem of \textcolor{black}{“Sort the blocks in the way you think it best fits”} with 4 bowls and 4 blocks, then our offline deployed ChatGLM2-6B model successfully generates the robot code policy \textcolor{black}{“response = ‘robot.pick\_and\_place(“pink block”, “yellow bowl”)\textbackslash n robot.pick\_and\_place(“pink block”, “pink bowl”)\textbackslash n robot.pick\_and\_place(“yellow block”, “pink bowl”)\textbackslash n robot.pick\_and\_place(“brown block”, “yellow bowl”)\textbackslash n robot.pick\_and\_place(“brown block”, “pink bowl”)\textbackslash n robot.pick\_and\_place(“blue block”, “yellow bowl”)\textbackslash n robot.pick\_and\_place(“blue block”, “pink’”} to successfully place the pink, brown, and blue blocks into the pink bowl, also put the yellow block into the yellow bowl within 2mins 42s via a 100\% task completion rate and 88\% performance score since the ViLD visual model has a few recognition errors for pink and yellow blocks. \textbf{Task B} focuses on the mission of \textcolor{black}{“Move the blocks to the corners”} with each 4 different color of bowls and blocks, then our ChatGLM2-6B model can generate the appropriate code action policy as presented in Fig. 3 to make the UR5 robotic arm fluently place the purple, brown, green, and blue blocks in the upper left corner of the desktop in 3mins 16s with the 100\% overall task completion rate and the overall performance score is 88\% since the robot execution process is lack of certain flexibility to just place all blocks uniformly in the upper left corner instead of placing them at different corners of the table desktop.  Transferred to \textbf{Task C} which aims to \textcolor{black}{“Move the blocks in the middle”}, our ChatGLM2-6B intelligently generates the corresponding robot code policy to make the UR5 robotic arm sequentially place the purple block and yellow block at the middle of table within 3mins 11s. Even though the task is completed with a whole 100\% rate, there exist redundant actions to move the yellow block which has been placed in the middle back to the left bowl during the execution process so that its final performance score is set to be 85\%. Based on the above experimental results, we obtain the overall clear accurate analysis that our innovatively designed “Socratic Models–ChatGLM” algorithm can successfully address and finish implementing diverse single instruction multi-step tasks including object color matching and moving objects to specific places (e.g., table corners or middle). In general, the proposed “Socratic Models–ChatGLM” can intelligently and quickly generate clear and right multi-step code policies via the locally deployed offline ChatGLM2-6B large language model for a new task scenario, then apply the CLIPort model to execute the generated action code for robotic control to complete multi-modal interactive tasks in a short period of execution time, achieving an overall performance score up to 85\%-90\%. In addition, it is shown that our “Socratic Models–ChatGLM” can further improve the performance by optimizing LLM code prompts as demonstrated in Fig. 2 or significant parameters related to code generation (e.g., max\_token, temperature) in LLMs for different multi-modal robotic interactive scenarios.

\subsection{Continuous Combination Multiple Task}
Furthermore, “Socratic Models-ChatGLM” even demonstrates outstanding performance in completing continuous combination multiple tasks via Fig. 4 and Fig. 5. Task A, B, C and D are continuous multi-task experiments conducted by user interactions within different multi-modal grasping scenario environments, and our locally deployed ChatGLM2-6B large language model can  generate code execution policies based on the user input text instruction and the visual information of the current desktop scenario detected by the ViLD model, then continuously complete 2, 3 or even 4 different multi-modal interactive control grasping tasks. 
\begin{figure}[htbp]
\vspace{-8pt}
\centerline{\includegraphics[width=0.486\textwidth]{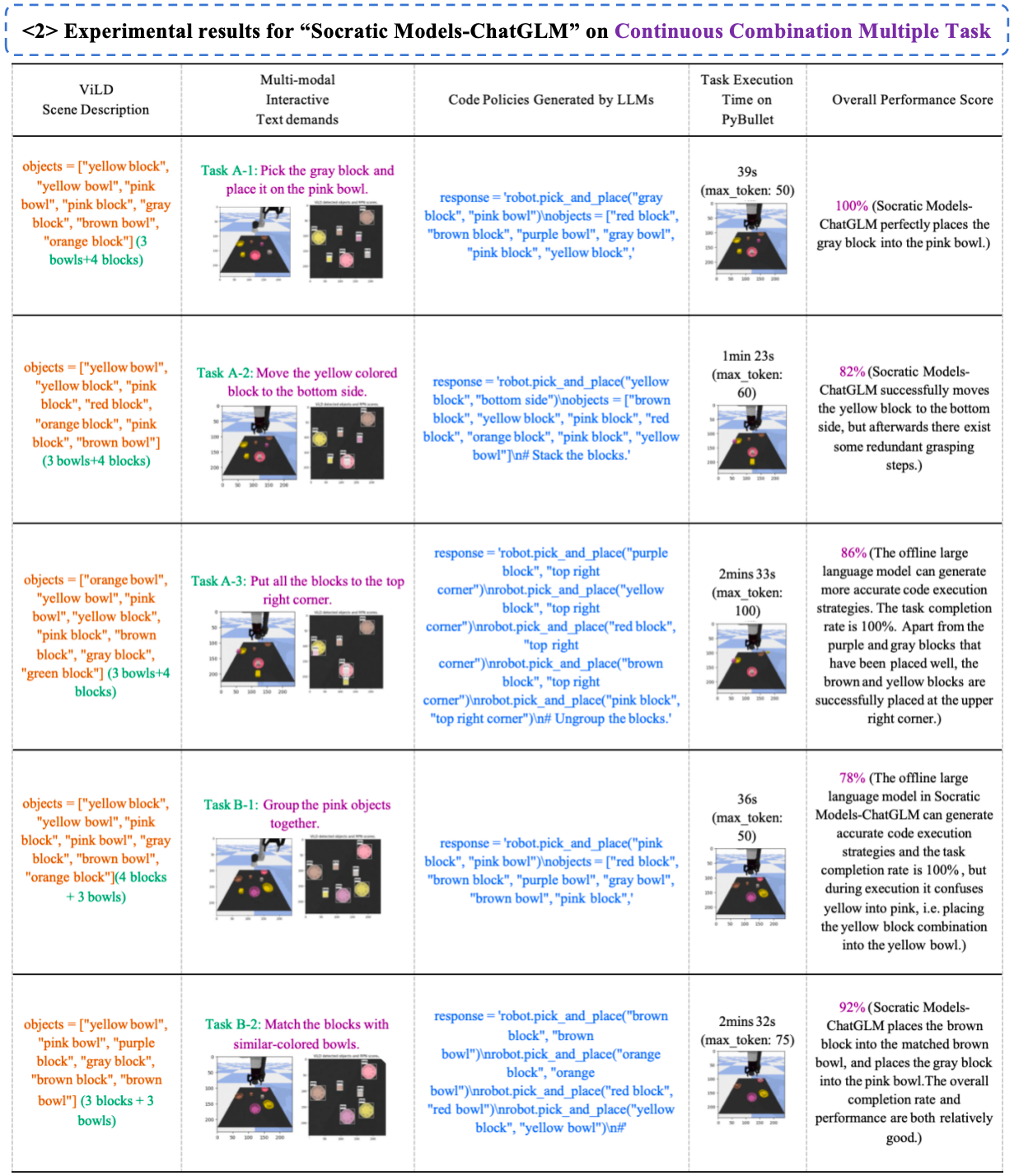}}
\caption{“Socratic Models-ChatGLM” Experiments on Continuous Combination Multiple Long-horizon Tasks (Part 1).}
\label{fig4}
\vspace{-18pt}
\end{figure}

Specifically \textbf{Task A} mainly involves three consecutive tasks for three bowls and four blocks of different colors in the initial multi-modal environment, in order: \ding{172} “Pick the gray block and place it on the pink bowl.” \ding{173} “Move the yellow colored block to the bottom side.” \ding{174} “Put all the blocks to the top right corner.”; After successfully picking the gray block and placing it on the pink bowl with ViLD, ChatGLM2-6B and CLIPort as shown in Fig. 1, we can continuously apply the ViLD algorithm for real-time visual reasoning to perform \textbf{Task A-2} in 1min 23s with the 82\% performance score to “move the yellow colored block to the bottom side”, then successfully conduct \textbf{Task A-3} “Put all the blocks to the top right corner” with a 100\% completion rate and 86\% performance score, which demonstrates that our “Socratic
Models-ChatGLM” can quickly and efficiently execute the
code strategy generated by the locally deployed ChatGLM2-6B offline model to present excellent potential in completing multi-step long-horizon tasks. In addition, we conduct more experiments on continuous combination multiple tasks 
\begin{figure}[htbp]
\centerline{\includegraphics[width=0.478\textwidth]{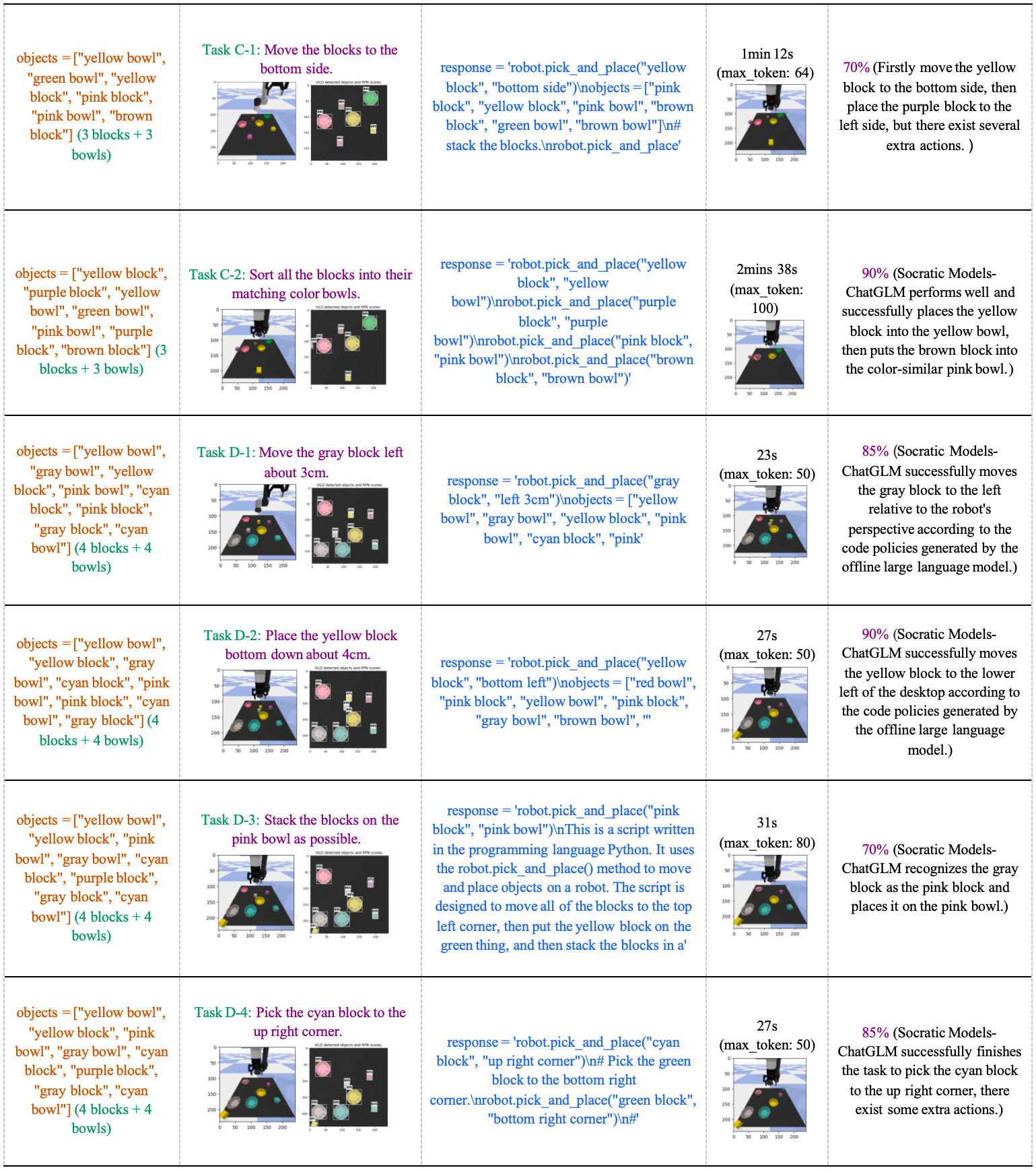}}
\caption{“Socratic Models-ChatGLM” Experiments on Continuous Combination Multiple Long-horizon Tasks (Part 2).}
\label{fig42}
\vspace{-17pt}
\end{figure}
\textbf{“Task B, C, D”} in Fig. 4 \& 5 with critical merits, even record the manipulation videos for each execution step to comprehensively evaluate “Socratic Models-ChatGLM”. Actually, \textbf{Task B} explores two interesting color-matching related tasks: \ding{172} Group the pink objects together. \ding{173} Match the blocks with similar-colored bowls. In general, our “Socratic Models-ChatGLM” can generate accurate code policies via the offline ChatGLM2-6B for the first mission and finish the second one with an excellent 100\% completion rate and 92\% good performance score. Next, \textbf{Task C} aims to firstly \ding{172} Move the blocks to the bottom side; and then \ding{173} Sort all the blocks into their matching color bowls, which motivates the offline LLMs to complete much more flexible multi-modal interactive robotic manipulation tasks without specific limiting standards via a better ViLD VLM detection performance. Furthermore, \textbf{Task D} successfully probes the original long-horizon task with four consecutive missions: \ding{172} Move the gray block left about 3cm; \ding{173} Place the yellow block bottom down about 4cm; \ding{174} Stack the blocks on the pink bowl as possible; \ding{175} Pick the cyan block to the up right corner. To emphasize, \textbf{Task D-1} and \textbf{Task D-2} creatively evaluate that the offline deployed LLMs can also achieve relatively equal excellent performance on moving a specific colored block at a certain distance to diverse orientations as the online OpenAI GPT-4 model in Code as Policies \cite{3}. Then our special “Socratic Models-ChatGLM” can even successfully finish the following block stacking and object placing interactive manipulation tasks, mainly benefited from the flexibility and unlimited reusability provided by our locally deployed ChatGLM large model, which exactly demonstrates much more outstanding potential on addressing complex multi-modal text-image multi-step long-horizon robotic manipulation tasks than the traditional Socratic Models \cite{4}. More detailed experimental result videos on Fig. 3--Fig. 5 can refer to \textcolor{black}{\url{https://github.com/2000222/Socratic-Models-ChatGLM}}.

\section{Conclusions}
In sum, we innovatively locally-deploy the offline LLM ChatGLM2-6B with OpenAI-compatible APIs via the facile FastChat platform and Autogen framework, then successfully integrate the visual and text information cleverly to implement the multi-modal interactive control algorithm “Socratic Models-ChatGLM”, which can not only quickly generate effective robot code action policies via the offline deployed ChatGLM2-6B model for novel interactive robotic scenarios, but also implement diverse single-instruction multi-step and complicated multi-modal continuous long-horizon manipulation tasks (e.g., object stacking, corner placement, item color matching) in a fast response time, achieving extraordinary performance scores up to 80\%-90\% and approximate 100\% task completion rates for better embodied interactions.

\addtolength{\textheight}{-12cm}   % This command serves to balance the column lengths

%%%%%%%%%%%%%%%%%%%%%%%%%%%%%%%%%%%%%%%%%%%%%%%%%%%%%%%%%%%%%%%%%%%%%%%%%%%%%%%%

%%%%%%%%%%%%%%%%%%%%%%%%%%%%%%%%%%%%%%%%%%%%%%%%%%%%%%%%%%%%%%%%%%%%%%%%%%%%%%%%

%%%%%%%%%%%%%%%%%%%%%%%%%%%%%%%%%%%%%%%%%%%%%%%%%%%%%%%%%%%%%%%%%%%%%%%%%%%%%%%%

%%%%%%%%%%%%%%%%%%%%%%%%%%%%%%%%%%%%%%%%%%%%%%%%%%%%%%%%%%%%%%%%%%%%%%%%%%%%%%%%

\end{document}